# Towards Machine Learning Prediction of Deep Brain Stimulation (DBS) Intra-operative Efficacy Maps

Camilo Bermudez*[a], William Rodriguez[b], Yuankai Huo[b], Allison E. Hainline[c], Rui Li[b], Robert Shults[b], Pierre D. D'Haese[b,d], Peter E. Konrad[d], Benoit M. Dawant[a,b], Bennett A. Landman[a,b]

[a] Department of Biomedical Engineering, Vanderbilt University, 2201 West End Ave, Nashville, TN, USA 37235;
[b] Department of Electrical Engineering, Vanderbilt University, 2201 West End Ave, Nashville, TN, USA 37235;
[c] Department of Biostatistics, Vanderbilt University, 2201 West End Ave, Nashville, TN, USA 37235;
[d] Department of Neurosurgery, Vanderbilt University Medical Center, 2201 West End Ave, Nashville, TN, USA 37235;

**ABSTRACT**

Deep brain stimulation (DBS) has the potential to improve the quality of life of people with a variety of neurological diseases. A key challenge in DBS is in the placement of a stimulation electrode in the anatomical location that maximizes efficacy and minimizes side effects. Pre-operative localization of the optimal stimulation zone can reduce surgical times and morbidity. Current methods of producing efficacy probability maps follow an anatomical guidance on magnetic resonance imaging (MRI) to identify the areas with the highest efficacy in a population. In this work, we propose to revisit this problem as a classification problem, where each voxel in the MRI is a sample informed by the surrounding anatomy. We use a patch-based convolutional neural network to classify a stimulation coordinate as having a positive reduction in symptoms during surgery. We use a cohort of 187 patients with a total of 2,869 stimulation coordinates, upon which 3D patches were extracted and associated with an efficacy score. We compare our results with a registration-based method of surgical planning. We show an improvement in the classification of intraoperative stimulation coordinates as a positive response in reduction of symptoms with AUC of 0.670 compared to a baseline registration-based approach, which achieves an AUC of 0.627 ($p < 0.01$). Although additional validation is needed, the proposed classification framework and deep learning method appear well-suited for improving pre-surgical planning and personalize treatment strategies.

## 1. INTRODUCTION

Deep brain stimulation (DBS) surgery is a neurosurgical procedure to alleviate symptoms in patients with movements disorders such as Parkinson's Disease (PD), essential tremor (ET), or primary dystonia [1]. During DBS surgery, a set of micro-electrodes is introduced into the brain parenchyma to test the effect of electrical stimulation at predetermined anatomical targets. Deep gray matter targets have been reported as having high efficacy for specific conditions, such as the subthalamic nucleus (STN) for PD, the ventral intermediate (VIM) nucleus of the thalamus for ET, or the globus pallidus interna (GPi) for dystonia [1]. The goal of the procedure is to find the anatomical location that maximizes efficacy in terms of symptomatic relief while minimizing the side effect profile of electrical stimulation to neighboring structures. An approximate target location is identified during presurgical planning and adjusted intraoperatively according to the stimulation profile of efficacy and side effects. A key challenge lies in identifying a presurgical target with a high likelihood of success, since accurate targeting can reduce surgical time and testing at multiple anatomical locations. Improved prediction may result in a diminishing intraoperative functional response from the patient and reduce medical morbidity.

Current methods of estimating the anatomical likelihood of stimulation efficacy use an image registration approach [2]. A probability map is created from intra-operative data of previous patients who have undergone the procedure. A profile of symptom relief and side effects is recorded at a number of intraoperative stimulation coordinates. The area affected by electrical stimulation is modeled as a sphere that represents the distribution of the electric field around the point of stimulation [2]. The radius of the sphere is a monotonic function of the current applied [2]. Previous work has used truncated Gaussians [3, 4] and annuli [2] as kernel functions for stimulation with similar results. Although symptom relief is a subjective measure (susceptible to patient variability and interpretation by the neurologist performing the exam), previous work has used a decrease in symptoms of over 50% intraoperatively as a robust measure of positive response across patients and examiners [2]. If a stimulation point has a positive response, it is mapped to a standard space, or atlas, using a non-rigid registration and modeled as a uniform probability distribution over the volume of the sphere. The positive response volumes from a population of patients are averaged to produce an efficacy probability map. The efficacy map is the registered to a new patient to identify the regions with high efficacy. Figure 1 shows the result of data acquisition during intraoperative stimulation in the STN. The blue regions represent the stimulated areas with a null response and the red regions the stimulated areas with a positive response. Each stimulated area is modelled as a sphere where the radius is a function of the current applied. These data show that only a few stimulation locations can be tested in each patient and are sparse with respect to the potential areas of interest.

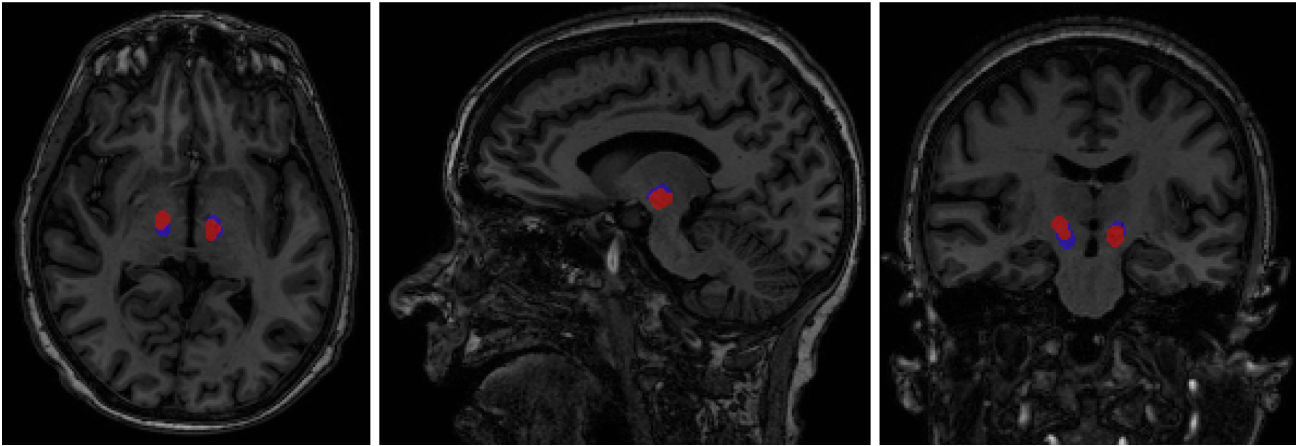

**Figure 1. Data acquisition process during DBS surgery in the subthalamic nucleus.** An example patient is shown with the efficacy data acquired intraoperatively. The region shown in blue represents all merged coordinates with the corresponding stimulation spheres that did not show a significant improvement in symptoms (ie. null response). The region shown in red are all merged coordinates with corresponding stimulation spheres with a significant reduction in symptoms (ie. positive response). The overlap between the two maps is resolved in favor of the positive response. This figure exemplifies the sparse nature of the data acquisition process and difficulty in visualizing underlying structures.

The current best practice method of producing efficacy probability maps follows anatomical guidance to identify the areas with the highest efficacy in a population. In this work, we propose to revisit this problem as a machine learning classification problem, where each voxel is a sample informed by the surrounding anatomy. We posit that the efficacy at each location can be learned from examples of intraoperative data to identify high efficacy areas without the use of a template space. We propose the use of a patch-based convolutional neural network to classify each patch centered at the stimulation coordinate as a positive response point. We present results for this classifier on 187 patients with a total of 2869 stimulation points and show that we can identify areas of high efficacy with higher accuracy than atlas-based methods.

## 2. METHODS

**2.1 Data Preparation**

A set of preoperative T1 magnetic resonance images (MRI) was acquired for each patient while they were anesthetized and their head was taped to the table to minimize motion under institutional review board approval. These images were acquired using a 3D SPGR sequence (TR 7.92 ms and TE 3.65 ms) in a 3T scanner. These MRI volumes consist of 256x256x170 voxels with a voxel resolution of 1x1x1 $mm^3$. DBS electrode implantation was then performed with a miniature stereotactic frame, the StarFix microTargeting Platform® (501(K), Number K003776, Feb. 23, 2001, FHC, INC; Bowdoin, ME). During surgery, a micropositioning drive (microTargeting® drive system, FHC Inc., Bowdoin, ME) was mounted on the platform. Recording and stimulating leads were then inserted using the microdrive. Details on the platform can be found in [5]. Intraoperative stimulation data used in this study were assessed by a clinical neurologist. For each stimulation coordinate, the efficacy was recorded as the percent decrease in symptoms. We also recorded current, side effects, and whether or not there was pass effect present. We used a total of 2869 stimulation points across 187 patients.

**2.2 Registration-based Approach**

We first replicate the current state of the art registration-based approach on the same dataset. First, we separate each stimulation coordinate as having a positive or null response with the protocol described above. Coordinates with a positive response are mapped to the Montreal Neurologic Institute (MNI-305) standard space using a non-rigid registration proposed by Rohde et al [6]. Once in standard space, the area affected by stimulation is modeled as a sphere centered at the stimulation coordinate with a radius proportional to the current applied [2]. The radius of the stimulation sphere was determined empirically from intraoperative measurements and recorded as a lookup table (Table 1). The radius for current values between those recorded on Table 1 was obtained by linear interpolation of known values. Each positive response stimulation volume is modeled as a uniform probability density function inside the sphere. We then average these probability density maps across all stimulation coordinates in standard space to obtain an efficacy probability map that can then be projected to a new test subject space.

**Table 1.** Lookup table for radius of stimulation sphere as a function of applied current.

| Current (mA) | Radius of stimulation sphere (mm) |
|---|---|
| < 1 | 1.00 |
| 1 | 1.80 |
| 2 | 2.42 |
| 3 | 2.94 |
| 4 | 3.33 |
| 5 | 3.72 |
| 6 | 4.05 |
| 7 | 4.35 |

In order to test the efficacy of these maps we perform a five-fold cross-validation scheme. For each fold, we use 80% of the subjects to generate the efficacy probability map in standard space and then project this to the remaining 20% of subjects to use as test subjects using a non-rigid registration [6]. This protocol was repeated a total of five times so that all subjects were used for testing only once. Once the probability efficacy map is projected to test subject space, we evaluate the probability of positive stimulation on the ground truth coordinates from intraoperative testing. First, we generate a sphere at each coordinate using the radius relationship in Table 1. The probability of classifying this coordinate as positive is the cumulative probability of the efficacy probability map within the stimulation sphere. Therefore, we obtain a probability estimate of classifying the intraoperative stimulation coordinate as positive or null using the population-based efficacy map registered from standard space. We assess the accuracy of this method with the receiver-operator characteristic (ROC) curve to find the best sensitivity and specificity that discriminates positive and null response spheres.

**2.3 Patch-based Classification**

In this work, we propose a patch-based neural network classifier to learn the function underlying the registration-based approach. The input data consists of two 3D volumes: first, the raw MRI of the patient and second, anatomical context provided by a volume with four labels: background/CSF, cortical gray matter, deep gray matter, and white matter. These labels were obtained using a multi-atlas approach where 45 atlases were non-rigidly registered [7] to the target image and non-local spatial staple (NLSS) label fusion [8] was used to fuse the labels from each atlas to the target image using the BrainCOLOR protocol [9].

The pipeline for this work is shown in Figure 2. First, we extract cubic patches from the raw T1 MRI and label map in subject space to use as input data. The patches are of size of size 51x51x51 mm and are centered at the intraoperative stimulation coordinate. A total of 2,869 3D patches were extracted. Each patch was then sorted into positive and null response groups. Positive response coordinates were selected in the same manner as the registration-based approach, where there must be greater than 50% improvement in symptoms during intraoperative stimulation. Exclusion criteria was the presence of pass effect in the intraoperative notes. We also select for the lowest current that maximizes efficacy at each coordinate. Of the 2,869 patches extracted, 34.37% were positive response coordinates while 65.63% were null response coordinates. The average current used in positive stimulation coordinates was 2.83 mA and the average current in null stimulation coordinates was 2.70 mA (p = 0.73, Wilcoxon-Mann-Whitney test).

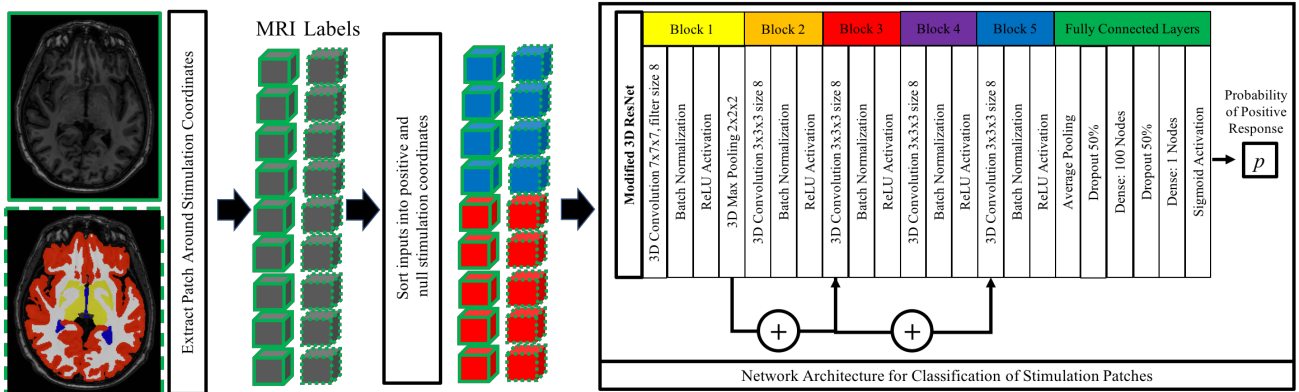

**Figure 2. Proposed pipeline for classification of response patches.** We use a modified 3D ResNet to train a classifier that discriminates input patches according to positive or negative intrasurgical response. The input data are two 3D patches centered around a stimulation coordinate: one is the raw MRI image and the other is a set of labels to provide anatomical context. The labels provided are background/CSF, cortical gray matter, deep gray matter, and white matter. The output of the network is the probability of the patch corresponding to a stimulation coordinate with a positive response.

A modified ResNet architecture [10] was used in this model to classify patches between positive and null responses (Figure 2). ResNet architectures have shown good performance on patch-based classification tasks [11-13]. The extracted patches were used as input and the class of positive or null response for the center of the patch was used as the output. The architecture is shown in Figure 2, which consists of a large convolution block followed by series of four convolutional blocks. All blocks consist of a 3D convolution layer, batch normalization, ReLU activation, and 3D max pooling of size 2x2x2 voxels. The initial convolution had a kernel of 7 voxels, while the following 4 convolutions had a kernel of 3 voxels. The first, third, and fifth convolutional blocks were joined via skip connections. After the convolutional blocks, there is a 3D average pooling followed by a fully connected dense layer of 100 nodes feeding into the final binary output classification with sigmoid activation. Dropout layers at 50% were used between the fully connected layers. The network was compiled using an Adam optimization algorithm using binary cross-entropy as the loss function with a learning rate of $1 \times 10^{-6}$ and a batch size of 32 patches.

Five separate models trained on 80% of the patients, using 10% of this subset as validation, with the remaining 20% as the testing set. A five-fold cross-validation scheme was used to assess the accuracy of the model on the testing set only. Results shown below reflect the accuracy of the testing set folds. All models were trained until the accuracy in the validation did not change by more than 5% over 10 epochs. Model accuracy was measured with the ROC curve to find the threshold with sensitivity and specificity. To assess the ability of the proposed method to generate efficacy maps at every voxel in the image, we generate a 3D patch centered at every voxel and use the classification network to compute the probability of the patch being classified as having a positive response.

## 3. RESULTS

### 3.1 Patch-Based Classification outperforms Registration-based accuracy

The proposed patch-based classification model shows an improvement in accuracy at discriminating positive and null response coordinates than the registration-based approach at the locations that we sampled. Figure 3 shows that the ROC curve plot for the proposed method and the registration-based method. The optimal cutoff was chosen by maximizing the true positive rate and minimizing the false positive rate with equal weighting. We show that the proposed method achieved an AUC of 0.670 with sensitivity of 0.302 and specificity of 0.885 at the point where the slope of the ROC curve becomes less than one. The registration-based approach achieved an AUC of 0.627 with a sensitivity of 0.338 and specificity of

0.849 at the optimal threshold. The proposed method shows a significant improvement in preoperative prediction of efficacy points compared to a registration-based approach (p<0.001, McNemar Test [14, 15]).

**3.2 Patch-based approach can generate patient-specific efficacy maps**

We assess the ability of the proposed method to generate a preoperative efficacy map across an entire brain volume. Figure 4 shows the efficacy map of a sample patient in the testing set overlaid on the MRI volume. We calculate the probability of a patch centered at that voxel being classified as having a positive response. The results show that the target areas predicted by our method are close to the positive response (green cross) and null response (blue ring) coordinates seen intraoperatively. There are many other areas of activation seen, particularly at border regions. It is important to note that all of the patches used to train the network were near the center of the image, where the structures of interest like the STN, GPi, and VIM are located.

## 4. DISCUSSION & CONCLUSION

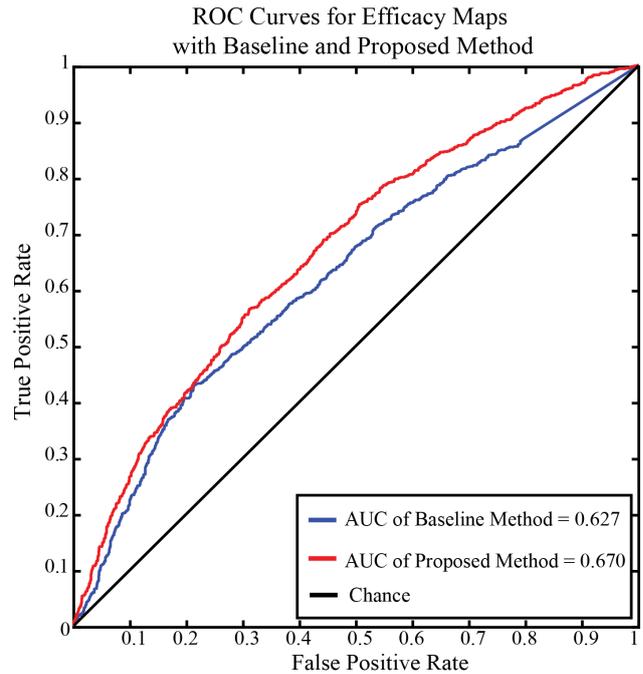

**Figure 3. ROC Curve for registration-based approach (baseline) and machine learning approach (proposed).** The AUC for the proposed method is 0.670 compared to an AUC of 0.627 using the registration-based approach (p<0.001, McNemar test).

Current state of the art in DBS preoperative planning uses a registration-based approach to produce an efficacy map that represents structures likely to respond to stimulation. This efficacy map serves as a surrogate for structures that are difficult to visualize from presurgical imaging such as the STN or the VIM. In this work, we propose a patch-based deep learning method to assist in preoperative planning for DBS surgery. We show that we can use data acquired from intraoperative recordings to learn the function between anatomic location and a positive response to stimulation at the points that have been sampled intra-operatively. Compared to a baseline registration-based approach, the proposed method can better predict the response to stimulation at the observed points when posed as a classification problem (Figure 3). However, it also leads to several infeasible responses with potential issues of symmetry and target localization (Figure 4). Note that the proposed classifier is a conditional classifier in that it is only valid in the domain upon which it was trained. The only data available are the points at which intraoperative recordings have been made, so no information is present at the cortex (and many other brain regions). So, the extrapolation seen in Figure 4 shows a high likelihood of positive response in the background and the cerebral cortex, which are outside the domain of stimulation of DBS surgery, likely due to similar intensity patterns. We include these results to demonstrate the differences in this tool from registration-based approaches, which are tied to the underlying anatomy and registration.

In this work, we see that we can better predict the response to stimulation in the observed coordinates. However, the proposed network is only valid when applied to regions that would have been sampled based on the clinical procedure. Hence, it is invalid to apply the classifier to areas such as the cortex. Further exploration is needed for the proposed machine learning approach to be ready for surgical consideration. For example, validation of the resulting efficacy maps could be achieved by comparing the efficacy probability against the final position of the electrode or even the active contact being used for symptomatic relief months after the surgery. The work presented here is a preliminary study that aims to demonstrate the feasibility of a machine learning approach to predict intraoperative clinical data. The clinical applicability of this work remains to be thoroughly tested and validated as described above before deploying to patient care.

The framework proposed herein introduces a new method to estimate efficacy maps during DBS presurgical mapping. This technique could potentially facilitate the development of accurate and patient-specific targets to improve functional outcomes during surgery. Moreover, an agnostic machine learning approach benefits from the lack of restrictions of registration and prespecified kernels, resulting in a personalized marker obtained from patient-specific anatomical context through imaging. As shown by the integration of segmentation labels, the proposed method is a natural framework to integrate additional contextual information, such as diffusion tensor imaging or clinical variables, which are difficult to include in the registration-based approach. These efforts will further refine and personalize the efficacy maps to discriminate between patient subpopulations, potential risk factors, and unique anatomical context. Further work should also include the side effects recorded during intraoperative stimulation to further refine the classification task to maximize efficacy while minimizing the side effect profile of stimulation, e.g., through multi-task learning.

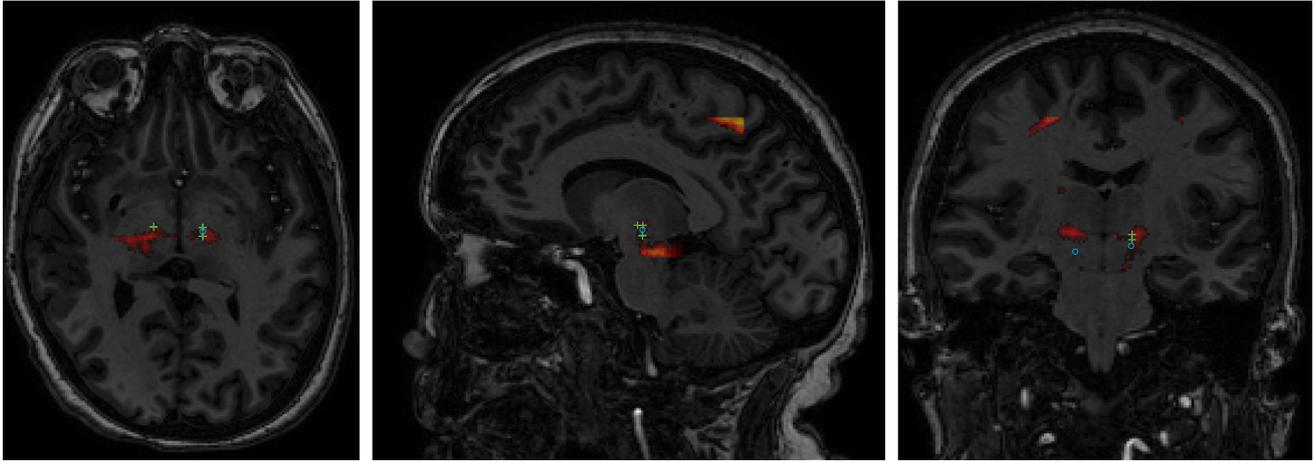

**Figure 4. Machine-learning approximation of an efficacy map.** Patient MRI image overlaid with probability of positive response assessed on a sliding patch basis (assuming an infeasible stimulation at every voxel). Green crosses indicate actual intraoperative stimulation locations with a positive response. Blue circles indicate intraoperative stimulation locations with a null response.

## ACKNOWLEDGEMENTS

This research was supported by NSF CAREER 1452485, NIH grants 1R03EB012461 (Landman), R01NS095291 (Dawant), and T32GM007347 (NIGMS/NIH). We are appreciative of the volunteers who gave their time and de-identified imaging data to be a part of this study. This study was in part using the resources of the Advanced Computing Center for Research and Education (ACCRE) at Vanderbilt University, Nashville, TN. This project was supported in part by ViSE/VICTR VR3029 and the National Center for Research Resources, Grant UL1 RR024975-01, and is now at the National Center for Advancing Translational Sciences, Grant 2 UL1 TR000445-06. This work does not reflect the opinions of the NIH or the NSF. We gratefully acknowledge the support of NVIDIA Corporation with the donation of the Titan X Pascal GPU used for this research.